\documentclass[runningheads]{llncs}
\usepackage[utf8]{inputenc}
\usepackage{kotex}  
\usepackage{multirow}
\usepackage{array}
\usepackage{amsmath}
\usepackage{placeins}

\usepackage{pifont}
\usepackage{makecell}
\usepackage[T1]{fontenc}
\usepackage{graphicx}
\usepackage[table]{xcolor}
\usepackage{xcolor}
\usepackage{bbding}
\usepackage{hyperref}
\hypersetup{pdfborder={0 0 0}}

\usepackage{url}
\begin{document}
\title{CoReVAD: A Contextual Reasoning Framework for Training-Free Video Anomaly Detection}

\titlerunning{CoReVAD}
%
\author{Hyeongmuk Lim\orcidID{0009-0007-9333-9124} \and Youngbum Hur\orcidID{0000-0002-1113-1730}\Envelope}
\authorrunning{H. Lim and Y. Hur}
%
\institute{{Department of Industrial Engineering, Inha University, Incheon, Republic of Korea} \\
\email{gudanr777@inha.edu}
\email{youngbum.hur@inha.ac.kr}}
\maketitle              
\begin{abstract}
Existing Video Anomaly Detection (VAD) methods typically rely on task-specific training, leading to strong domain dependency and high training costs. Moreover, most existing methods output only scalar anomaly scores, providing limited insight into why specific events are considered abnormal. Recent advances in Vision–Language Models (VLMs) have enabled both anomaly detection and human-interpretable reasoning. However, many VLM-based approaches still require additional training steps (e.g., instruction tuning or verbalized learning) or external Large Language Models (LLMs), incurring further training costs and inference overhead. To address these challenges, we propose CoReVAD, a contextual reasoning framework for training-free video anomaly detection that operates with a single frozen VLM. CoReVAD directly generates anomaly scores and temporal descriptions from the VLM. To mitigate noise in generative outputs, we introduce a Local Response Cleaning (LRC) module based on local vision–text alignment. Furthermore, global temporal context and progression are incorporated through softmax-based refinement, Gaussian smoothing, and position weighting. Experiments on UCF-Crime and XD-Violence demonstrate that CoReVAD achieves competitive performance among training-free methods while providing reliable and interpretable explanations. Our official code is available at: \url{https://github.com/Muk-00/CoReVAD}
\end{abstract}
\begin{figure}
\centering
\includegraphics[width=\textwidth]{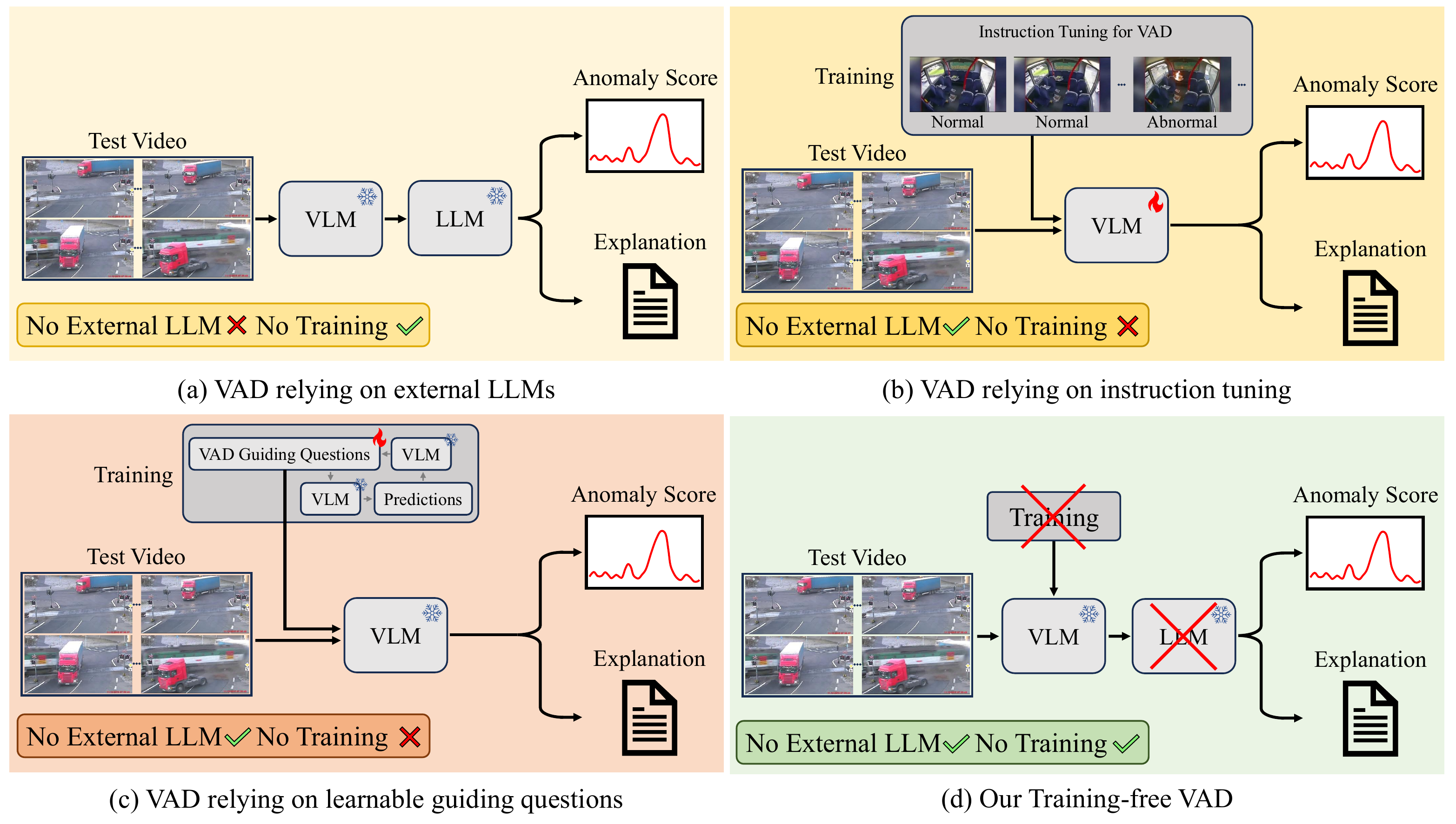}
\caption{Existing explainable VAD methods can be broadly categorized into three types: (a) training-free VAD methods that rely on external LLMs for reasoning, (b) VAD methods that adapt VLMs via instruction tuning and (c) VAD methods that introduce learnable guiding questions through verbalized learning. In contrast, (d) Our CoReVAD proposes an explainable VAD framework that requires neither external LLMs nor any training phase.}
\label{fig1}
\end{figure}
\section{Introduction}
Video Anomaly Detection (VAD) aims to identify abnormal events within a video sequence. It has become an essential technology in industrial applications, including public safety and autonomous driving \cite{bogdoll2022anomaly,zhu2020video}. The extensive spatio-temporal complexity of video data makes large-scale annotation highly costly. Although VAD has advanced substantially across a variety of approaches, several fundamental challenges remain unresolved.
First, traditional VAD models inherently require a training phase, which makes them domain-specific and limits generalization to unseen environments. Second, existing methods output only anomaly scores without providing human-interpretable reasoning, making it difficult to understand why certain frames are identified as abnormal. Recent advances in Vision-Language Models (VLMs) \cite{chen2024internvl,liu2023visual} have introduced new possibilities for addressing these challenges. 
VLMs combine language-based reasoning with the visual understanding provided by pre-trained vision encoders, enabling them to describe visual scenes in natural language and infer anomalous behaviors. As shown in Fig.~\ref{fig1}, recent studies have proposed a variety of approaches for leveraging off-the-shelf frozen VLMs in VAD. Some approaches apply instruction tuning to align VLMs better with anomaly detection tasks \cite{lv2024video,zhang2025holmes}. Although this can substantially improve both anomaly detection and reasoning performance, instruction tuning requires additional datasets and training costs, making real-world deployment challenging. Other methods forward VLM-produced scene descriptions to external Large Language Models (LLMs) \cite{achiam2023gpt,touvron2023llama} for scoring anomalies or generating explanations \cite{dev2024mcanet,yang2024follow,zanella2024harnessing}. While these methods benefit from the strong reasoning capability of LLMs, they lead to significant computational overhead because the two models operate in separate stages. Driven by these developments, recent research has explored VAD methods using verbalized learning \cite{xiao2024verbalized}. VERA \cite{ye2025vera} learns guiding prompts through a verbalized learning framework. Although it does not require an additional LLM or fine-tuning, it still depends on an external training phase to obtain guiding prompts, thereby increasing domain dependency. To address these challenges, we introduce {\bfseries CoReVAD}, a {\bfseries Co}ntextual {\bfseries Re}asoning framework for training-free {\bfseries VAD}, which operates with a single frozen VLM. CoReVAD generates segment-level responses including anomaly decisions and temporal descriptions from the VLM without requiring external LLMs, thereby reducing inference complexity. However, VLMs are often noisy and produce inaccurate outputs. We address this issue by introducing a Local Response Cleaning (LRC) module that leverages textual outputs around segments to refine VLM-derived responses by estimating how well they match the visual content (i.e., vision–text alignment). Finally, to account for global temporal context—capturing the overall temporal relationships across the entire video—we apply a softmax-based refinement to the segment-level binary anomaly scores. We further adjust the temporal progression of anomalies through Gaussian smoothing and position weighting. CoReVAD can be applied across domains without any additional training. We demonstrate that CoReVAD outperforms one-class, unsupervised, and training-free VAD methods on UCF-Crime \cite{sultani2018real} and XD-Violence \cite{wu2020not}.
In summary, the key contributions of this paper are as follows:
\begin{enumerate}
    \item We introduce CoReVAD, a training-free video anomaly detection framework based on a single pre-trained VLM that performs anomaly detection and provides human-interpretable reasoning, without requiring domain knowledge, external LLMs, or additional data collection.
    \item We propose a contextual refinement strategy that suppresses noise in VLM responses through Local Response Cleaning (LRC) and refines anomaly scores using global cosine-similarity-based softmax weighting, enabling stable frame-level detection.
    \item Experiments demonstrate that our method achieves competitive performance compared with training-free methods and remains comparable to training-based VAD methods.

\end{enumerate}
\section{Related work}
\subsection{Video Anomaly Detection}
Video Anomaly Detection (VAD) focuses on identifying abnormal events at the frame-level within a given video. Since providing frame-level annotations for videos requires substantial manual effort and annotation cost, many studies have attempted to address this limitation by adopting different levels of supervision (e.g., unsupervised, weakly supervised). Unsupervised VAD methods (often referred to as one-class classification (OCC)) \cite{gong2019memorizing,liu2018future,liu2021hybrid,park2020learning,zaheer2022generative} learn normal patterns and detect anomalies by measuring reconstruction or prediction inconsistencies. Lee et al. \cite{lee2022multi} improve VAD performance by building a multi-contextual prediction framework based on Vision Transformers (ViT) \cite{dosovitskiy2020image}. Similarly, Zhao et al. \cite{zhao2022exploiting} introduce a prediction-based model that captures spatio-temporal dependencies for anomaly detection. Several studies have proposed fully unsupervised learning without labels for either normal or abnormal videos \cite{thakare2023rareanom,thakare2023dyannet,zaheer2022generative}. Although such methods benefit from extremely low annotation cost, the absence of explicit supervision makes it challenging for them to achieve high detection performance. On the other hand, weakly supervised VAD (WSVAD) methods leverage video-level annotations to learn discriminative characteristics of normal and abnormal events and therefore typically achieve high detection performance \cite{li2022scale,li2022self,sultani2018real,wu2022self,wu2021learning}. Most WSVAD frameworks employ Multiple Instance Learning (MIL), first introduced by Sultani et al. \cite{sultani2018real}, where a video is treated as a bag containing multiple segments as instances. Tian et al. \cite{tian2021weakly} improve MIL-based VAD by incorporating feature magnitude learning and leveraging self-attention mechanisms to capture both short- and long-range temporal dependencies. Recent works have also explored the use of CLIP \cite{radford2021learning} for VAD \cite{joo2023clip,wu2024vadclip}. In particular, Wu et al. \cite{wu2024vadclip} achieve fine-grained anomaly detection by utilizing both visual features and text embeddings derived from video labels. Despite the strong performance of these traditional VAD approaches, they inherently require training, which leads to domain dependency. Moreover, these methods rely on score-based outputs, which provide limited insight into why specific frames are classified as anomalous.
\subsection{Vision-Language Models in VAD} 
The recent remarkable success of VLMs has prompted increasing interest in applying them to VAD \cite{lv2024video,shao2025eventvad,yang2024follow,zanella2024harnessing,zhang2025holmes}. These approaches attempt to improve the limited explainability inherent in traditional VAD methods. Zhang et al. \cite{zhang2025holmes} construct a large-scale hierarchical VAD dataset and perform instruction tuning to achieve strong detection accuracy and anomaly reasoning. However, this approach requires substantial data collection, exhibits domain dependency, and involves additional training procedures. Zanella et al. \cite{zanella2024harnessing} first propose a training-free method by generating captions with a VLM and using pre-trained LLMs for anomaly scoring, and Dev et al. \cite{dev2024mcanet} further extend this idea by integrating image, audio, and text modalities. Although these approaches can be applied across domains without additional training, Ye et al. \cite{ye2025vera} report that decoupling a VAD system into a frozen VLM and an external LLM introduces considerable inference overhead. In contrast, we propose CoReVAD, a training-free VAD framework that relies solely on a single pre-trained VLM, avoiding both instruction tuning and external LLMs. CoReVAD enables efficient inference without additional overhead and avoids domain-specific dependence, achieving stable and competitive performance compared to existing training-free methods.

\section{Methodology}
\subsection{Problem Formulation}
\label{sec:3.1}
Consider a video $V$ consisting of $F$ frames, where $I_i$ denotes the $i$-th frame of $V$. In standard labeling, anomalous frames are labeled as 1 and normal frames as 0, so the ground-truth label sequence for $V$ is defined as $Y = [y_1,\dots,y_F]$ with $y_i \in \{0,1\}$. Existing VAD methods estimate anomaly scores by employing various techniques. However, these approaches typically require large-scale training datasets and annotations, making them difficult to deploy in real-world scenarios. To overcome these limitations, we propose CoReVAD, which estimates anomaly scores using only a pre-trained VLM $\mathrm{\Phi}_\text{VLM}$ and multimodal encoders $f_\text{vision}$ and $f_\text{text}$, without any additional training. For a video $V$, the predicted anomaly score sequence is represented as $\hat{Y} = [\hat{y}_1,\dots,\hat{y}_F]$ with $\hat{y}_i \in \{0,1\}$. The proposed framework enables detecting anomalous events and generating interpretable descriptions by incorporating visual and textual contexts.
\subsection{VLM-Based Anomaly Detection and Reasoning Generation}
\begin{figure}
\centering
\includegraphics[width=\textwidth]{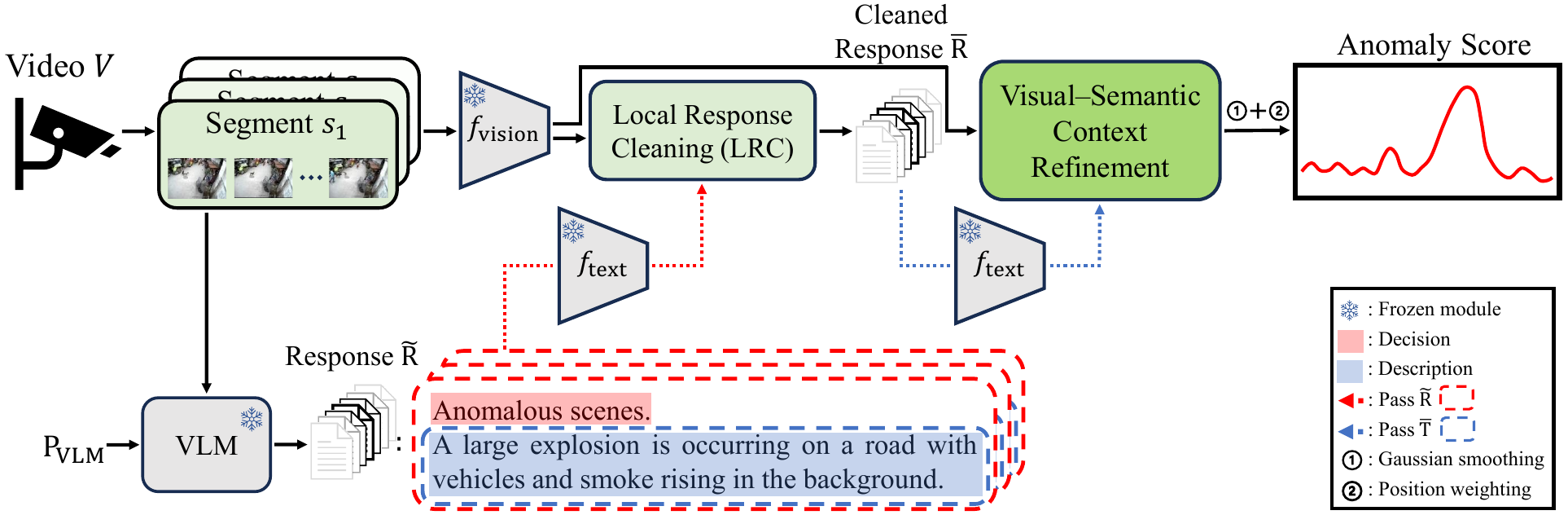}
\caption{Architecture of our CoReVAD. The pre-trained VLM first generates a response $\tilde{r}_j$ (red bounding box) for each segment $s_j \in V$, including a segment-level anomaly decision and a temporal description (blue bounding box), forming a response sequence $\tilde{R}$. The Local Response Cleaning (LRC) refines $\tilde{r}_j$ by selecting the most semantically aligned one response from neighboring responses using $f_\text{vision}$ and $f_\text{text}$. Visual–Semantic Context Refinement then leverages refined descriptions by applying a softmax-based similarity to incorporate global temporal context into the segment-level anomaly scores. Finally, Gaussian smoothing and position weighting are applied to obtain precise anomaly scores.}
\label{fig2}
\end{figure}
\begin{figure}
\centering
\includegraphics[width=\textwidth]{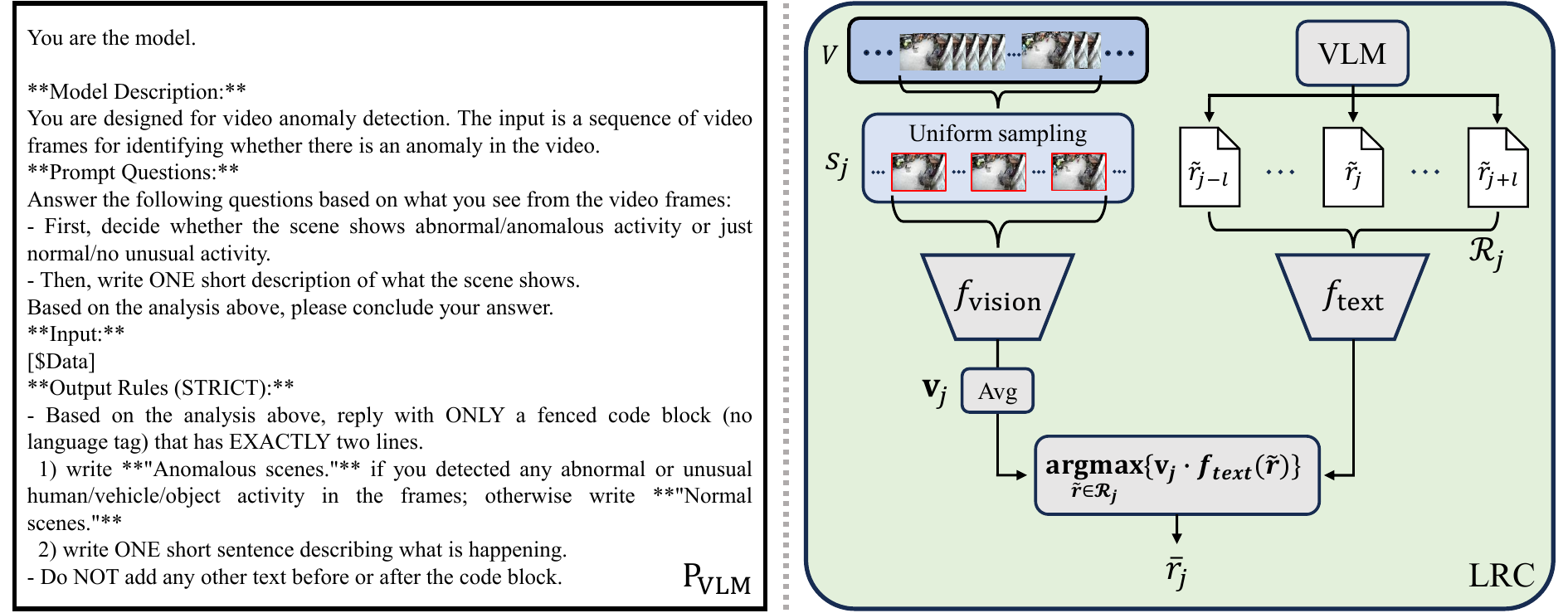}
\caption{$\text{P}_\text{VLM}$ and LRC.}
\label{fig3}
\end{figure}

\label{sec:3.2}
\noindent{An overview of our framework is shown in Fig.~\ref{fig2}. We generate initial binary decisions and interpretable descriptions using the VLM. However, directly feeding every frame into the VLM incurs substantial computational cost. To address this issue, we divide a video $V$, consisting of $F$ frames, into segments $S = [s_1,\dots,s_M]$ at a fixed interval $d$, where  $M = \lfloor F/d \rfloor + 1$ denotes the total number of segments. For each segment, we uniformly sample $n$ frames and feed the sampled frames, together with a predefined prompt $\text{P}_\text{VLM}$ (see Fig.~\ref{fig3}), into the VLM to obtain segment-level responses $\tilde{R}=[\tilde{r}_1, \dots, \tilde{r}_M]$. Formally, the VLM response $\tilde{R}$ is defined as:
\begin{equation}
\label{equ1}
    \tilde{R} = \mathrm{\Phi}_\text{VLM}(S;\mathrm{P}_\text{VLM})
\end{equation}
where each response $\tilde{r}_j \in \tilde{R}$ is converted to derive a binary anomaly score $\tilde{y}_j$, such that $\tilde{y}=1$ if $\tilde{r}_j$ includes “Anomalous scenes” and $\tilde{y}_j = 0$ if it includes "Normal scenes". Accordingly, the response set $\tilde{R}$ produces segment-level anomaly scores $\tilde{Y}=[\tilde{y}_1, \dots, \tilde{y}_M]$ and temporal descriptions $\tilde{T} = [\tilde{t}_1, \ldots, \tilde{t}_M]$ (i.e., the textual components of $\tilde{R}$, excluding the anomaly decisions), which jointly provides reasoning that reflects both visual and temporal information.}

\noindent {\bfseries Local Response Cleaning.} The raw response $\tilde{R}$ may contain noise due to randomness in generative outputs, leading to incorrect anomaly decisions or irrelevant explanations. Meanwhile, adjacent video frames typically share similar visual semantics because a video is typically recorded at high frame rates. Motivated by this observation, we introduce a local vision-text alignment mechanism for response cleaning. Prior works \cite{dev2024mcanet,zanella2024harnessing} perform cleaning over the entire video sequence, which may incorporate information from distant scenes that are not semantically related. As a result, the cleaning process can be adversely affected by irrelevant context, leading to incorrect explanations and degraded anomaly detection performance.
In contrast, we propose Local Response Cleaning (LRC), which treats cleaning as finding the semantically closest response to a target response $\tilde{r}_j$ from a local neighborhood, rather than across the entire video. As illustrated in Fig.~\ref{fig3}, LRC leverages the local response set $\mathcal{R}_j$ to account for local temporal context and obtain an accurate response $\bar{r}_j$ as:
\begin{equation}
\label{equ2}
    \bar{r}_j = \operatorname*{argmax}_{\tilde{r} \in \mathcal{R}_j}\{\mathrm{v}_j \cdot f_\text{text}(\tilde{r})\}
\end{equation}
where $\{ \cdot \}$ denotes the cosine similarity. Here, $\mathrm{v}_j$ represents the average vision feature extracted from the uniformly sampled frames of segment $s_j$ using $f_\text{vision}$. The text encoder $f_\text{text}$ encodes the response $\tilde{r} \in \mathcal{R}_j$ into the joint vision--text embedding space. $\mathcal{R}_j = [\tilde{r}_{j-l},\dots,\tilde{r}_{j+l}]$ denotes the local response set centered at segment $j$, which consists of responses from $l$ temporally neighboring segments on each side. Through this process, we obtain the cleaned segment-level anomaly scores $\bar{Y}$ and temporal descriptions $\bar{T}$.
\subsection{Visual and Temporal Context Refinement}
\label{sec:3.3}
{\bfseries Visual--Semantic Context Refinement.} The anomaly scores obtained in Sect.~\ref{sec:3.2} have two main challenges. First, VLM outputs rely on short-range temporal context and thus fail to capture long-range temporal context. Second, the obtained scores fail to capture temporal dynamics and progression within a video. To overcome the first challenge, we incorporate global temporal context by combining visual and semantic similarity with anomaly scores across the entire video sequence. We refine the segment-level score $\rho_j$ as:
\begin{equation}
\label{equ3}
    \rho_j=\sum_{i=1}^M\bar{y}_j \cdot\frac{\exp{(\{\mathbf{v}_j\cdot f_\text{text}(\bar{t}_i)\}}/\tau)}{\sum_{i=1}^M\exp{(\{\mathbf{v}_j\cdot f_\text{text}(\bar{t}_i)\}}/\tau)}
\end{equation}
where $\{ \cdot \}$ and $\mathrm{v}_j$ follow Eq.~\ref{equ2}. Here $f_\text{text}(\bar{t}_i)$ encodes the cleaned temporal description $\bar{t}_j \in \bar{T}$ from LRC. The vision-text similarities are normalized across all temporal descriptions using a softmax function with temperature hyperparameter $\tau$, thereby refining the cleaned segment-level anomaly score $\bar{y}_j$. This refinement produces the segment-level anomaly scores $\rho$ that capture global scene and semantic context as well as local temporal context. \\
\noindent{{\bfseries Temporal Dynamics and Progression Refinement.} Then, we refine the segment-level scores into frame-level scores by integrating temporal progression.} In real-world scenarios, anomalous behaviors rarely appear suddenly. Instead, they tend to grow more severe over time. Therefore, we aim to obtain stable anomaly scores by reflecting temporal continuity. To achieve this, we apply Gaussian smoothing to obtain a continuous representation of segment-level scores.  We obtain the segment-level anomaly scores as: 
\begin{equation}
\label{equ4}
    \Gamma = \rho \ast G = [\gamma_{1},\dots,\gamma_{M}]. 
\end{equation}
where $\ast$ denotes a 1D convolution operation and the Gaussian kernel is given by $G(p)=\exp(\frac{-p^2}{2\sigma_1^2})$, with $p$ representing the distance from the kernel center and $\sigma_1$ the variance. After Gaussian smoothing, we assign $\gamma_j$ to the frames within the $j$-th segment  to obtain the frame-level anomaly scores $[\hat{\gamma}_1,\dots,\hat{\gamma}_F]$. Finally, to reflect the tendency of abnormal patterns in videos, we apply a Gaussian function as position weights $w(i)$ and refine the final frame-level anomaly scores as follows:
\begin{equation}
\label{equ5}
    \hat{y}_i = w(i) \cdot \hat{\gamma}_i
\end{equation}
where $w(i) = \exp(\frac{-(i-c)^2}{2\sigma_2}), (1 \leq i \leq F)$, $c = \lfloor \frac{F}{2} \rfloor$ is the center frame index, and $\sigma_2$ is the variance. By combining these factors, we obtain precise frame-level anomaly scores $\hat{Y} = [\hat{y}_{1},\dots,\hat{y}_{F}]$.
\section{Experiments}
In this section, we compare our method with existing approaches that are trained with different levels of supervision using two benchmark and evaluation metrics. Sect.~\ref{sec:4.1} describes the datasets and metrics used for evaluation, and Sect.~\ref{sec:4.2} outlines the implementation details. Sect.~\ref{sec:4.3} presents quantitative comparisons with state-of-the-art VAD methods, and Sect.~\ref{sec:4.4} investigates the contribution of each component in our framework. Finally, Sect.~\ref{sec:4.5} provides qualitative results that demonstrate the effectiveness of our method.
\begin{table}
\renewcommand{\arraystretch}{1.1}
\centering
\caption{Comparison with state-of-the-art VAD methods on UCF-Crime.}
\label{tab1}

\begin{tabular}{
  >{\hspace{2pt}}p{2.5cm} |
  >{\hspace{2pt}}p{2.1cm} |
  >{\hspace{2pt}}p{4.4cm} |
  >{\centering\arraybackslash}p{1.4cm}
}
\hline
\textbf{Explainable} & \textbf{Category} & \textbf{Method} & \textbf{AUC(\%)}\\
\cline{1-4}

\multirow{16}{*}{\shortstack{Non-explainable\\VAD Methods}}
    & \multirow{9}{*}{\shortstack[l]{Weakly\\supervised}}
     & Sultani \cite{sultani2018real} & 77.92 \\
    & & GCL \cite{zaheer2022generative} & 79.84 \\
    & & Wu et al. \cite{wu2020not} & 82.44 \\
    & & RTFM \cite{tian2021weakly} & 84.30 \\
    & & MSL \cite{li2022self} & 85.62 \\
    & & S3R \cite{wu2022self} & 85.99 \\
    & & MGFN \cite{chen2023mgfn} & 86.98 \\
    & & CLIP-TSA \cite{joo2023clip} & 87.58 \\
    & & VADCLIP \cite{wu2024vadclip} & 88.02 \\
\cline{2-4}

    & \multirow{5}{*}{\centering   One-class}
     & SVM \cite{sultani2018real} & 50.00 \\
    & & Hasan et al. \cite{hasan2016learning} & 51.20 \\
    & & SSV \cite{sohrab2018subspace} & 58.50 \\
    & & BODS \cite{wang2019gods} & 68.26 \\
    & & GODS \cite{wang2019gods} & 70.46 \\
\cline{2-4}

    & \multirow{2}{*}{\centering   Unsupervised}
     & GCL \cite{zaheer2022generative} &  71.04 \\
    & & DyAnNet \cite{thakare2023dyannet} & 79.76 \\
\hline

\multirow{11}{*}{\shortstack{Explainable\\VAD Methods}}
    & \multirow{2}{*}{\centering   Fine-tuning}
      & VADor \cite{lv2024video} & 88.10 \\
    & & Holmes-VAU \cite{zhang2025holmes} & 88.96 \\
\cline{2-4}
    & \multirow{1}{*}{\centering Verbalized}
      & VERA \cite{ye2025vera} & 86.55 \\
\cline{2-4}
    & \multirow{8}{*}{\centering Training-free}
      & ZS CLIP \cite{zanella2024harnessing} & 53.16 \\
    & & ZS IMAGEBIND (Image) \cite{zanella2024harnessing} & 53.65 \\
    & & ZS IMAGEBIND (Video) \cite{zanella2024harnessing} & 55.78 \\
    & & LLAVA-1.5 \cite{gu2024anomalygpt} & 72.84 \\
    & & LAVAD \cite{zanella2024harnessing} & 80.28 \\
    & & EventVAD \cite{shao2025eventvad} & 82.03 \\
    & & MCANet \cite{dev2024mcanet} & 82.47 \\
    & & \textbf{Ours} & \textbf{82.51} \\
\hline

\end{tabular}
\end{table}
\FloatBarrier
\subsection{Experimental Setting}
\label{sec:4.1}
{\bfseries Datasets.}  We evaluate our method using two large-scale VAD benchmarks, UCF-Crime \cite{sultani2018real} and XD-Violence \cite{wu2020not}. {\bfseries UCF-Crime} consists of 1,900 untrimmed real-world surveillance videos with 13 distinct anomaly behaviors (e.g., abuse, fighting) and 290 test videos. {\bfseries XD-Violence} consists of 4,754 untrimmed videos sourced from both movies and YouTube. It includes 6 distinct anomaly behaviors (e.g., car accidents, riots) and 800 test videos. {\bfseries Metric.} Following previous works \cite{dev2024mcanet,lv2024video,zhang2025holmes}, we measure the VAD performance using the area under the curve (AUC) of the frame-level receiver operating characteristics (ROC). For XD-Violence, we additionally measure the average precision (AP) following \cite{wu2020not}.
\subsection{Implementation Details}
\label{sec:4.2}
We employ InternVL2-8B \cite{chen2024internvl} as $\mathrm{\Phi}_\text{VLM}$ to generate anomaly scores and descriptions. Each video $V$ is divided into segments of $d=30$ frames, and for each segment we uniformly sample $n=8$ frames as input to the $\mathrm{\Phi}_\text{VLM}$. For LRC and Score Refinement, we use the CLIP vision encoder (ViT-B/16) as $f_\text{vision}$ and the text encoder as $f_\text{text}$. In LRC, we use a neighboring context length of $l=1$ to compare with adjacent responses. We also use temperature $\tau=0.05$ and apply Gaussian smoothing with $p=9$ and $\sigma_1=5$ for smooth temporal refinement. For position weighting, we set $\sigma_2=\lfloor\frac{F}{2}\rfloor$, where $F$ denotes the number of frames.
\begin{table}
\renewcommand{\arraystretch}{1.05}
\centering
\caption{Comparison with state-of-the-art VAD methods on XD-Violence.}
\label{tab2}
\begin{tabular}{
  >{\hspace{2pt}}p{2.5cm} |
  >{\hspace{2pt}}p{2.1cm} |
  >{\hspace{2pt}}p{4.4cm} |
  >{\centering\arraybackslash}p{1.1cm}
  >{\centering\arraybackslash}p{1.4cm}
}
\hline
\textbf{Explainable} & \textbf{Category} & \textbf{Method} & \textbf{AP(\%)} & \textbf{AUC(\%)}\\
\cline{1-5}

\multirow{13}{*}{\shortstack{Non-explainable\\VAD Methods}}\
    & \multirow{8}{*}{\shortstack[l]{Weakly\\supervised}}
      & Wu et al. \cite{wu2020not} & 73.20 & - \\
     & & Wu \& Liu \cite{wu2021learning} & 75.90 & - \\
     & & RTFM \cite{tian2021weakly} & 77.81 & - \\  
     & & MSL \cite{li2022self} & 78.58 & - \\
     & & S3R \cite{wu2022self} & 80.26 & - \\
     & & MGFN \cite{chen2023mgfn} & 80.11 & - \\
     & & CLIP-TSA \cite{joo2023clip} & 82.17 & - \\
     & & VADCLIP\cite{wu2024vadclip} & 84.51 & - \\
\cline{2-5}
    & \multirow{4}{*}{\centering One-class}
      & Hasan et al. \cite{hasan2016learning} & - & 50.32 \\
     & & Lu et al. \cite{lu2013abnormal} & - & 53.56 \\
     & & BODS \cite{wang2019gods} & - & 57.32 \\
     & & GODS \cite{wang2019gods} & - & 61.56 \\
\cline{2-5}
    & \multirow{1}{*}{\centering Unsupervised}
      & RareAnom \cite{thakare2023rareanom} & - & 68.33 \\
\hline

\multirow{11}{*}{\shortstack{Explainable\\VAD Methods}}
    & \multirow{1}{*}{\centering Fine-tuning}
      & Holmes-VAU\cite{zhang2025holmes} & 87.68 & - \\
\cline{2-5}
    & \multirow{1}{*}{\centering Verbalized}
      & VERA \cite{ye2025vera} & 70.54 & 88.26 \\
\cline{2-5}
    & \multirow{8}{*}{\centering Training-free}
      & ZS CLIP \cite{zanella2024harnessing} & 17.83 & 38.21 \\
     & & ZS IMAGEBIND (Image) \cite{zanella2024harnessing} & 27.25 & 58.81 \\
     & & ZS IMAGEBIND (Video) \cite{zanella2024harnessing} & 25.36 & 55.06 \\
     & & LLAVA-1.5 \cite{gu2024anomalygpt} & 50.26 & 79.62 \\
     & & LAVAD \cite{zanella2024harnessing} & 62.01 & 85.36 \\
      & & EventVAD \cite{shao2025eventvad} & 64.04 & 87.51 \\
     & & MCANet \cite{dev2024mcanet} & 69.72 & 87.43 \\ 
     & & \textbf{Ours} & \textbf{70.94} & \textbf{91.44} \\
\hline

\end{tabular}
\end{table}
\FloatBarrier
\subsection{Comparison with State-of-the-Art Methods}
\label{sec:4.3}

In this section, we compare CoReVAD with state-of-the-art VAD methods, including both non-explainable and explainable approaches. Table~\ref{tab1} presents the results on UCF-Crime. CoReVAD achieves the highest AUC ROC among training-free methods  \cite{dev2024mcanet,gu2024anomalygpt,zanella2024harnessing}. Moreover, it outperforms one-class classification methods \cite{hasan2016learning,sohrab2018subspace,sultani2018real,wang2019gods} as well as unsupervised methods \cite{thakare2023dyannet,zaheer2022generative}. Given that our proposed method is training-free, outperforming training-based methods is highly significant. Specifically, it surpasses LAVAD \cite{zanella2024harnessing} by 2.23\%. We also report the comparison with state-of-the-art methods on XD-Violence in Table~\ref{tab2}. Compared to MCANet \cite{dev2024mcanet}, CoReVAD improves AP by 1.22\% and AUC ROC by 4.01\%. Moreover, CoReVAD surpasses VERA \cite{ye2025vera}, which represents the state-of-the-art explainable VAD method, by 3.18\% in AUC ROC. We acknowledge that fine-tuning methods \cite{lv2024video,zhang2025holmes} achieve superior performance, and that VERA \cite{ye2025vera} reports superior results on UCF-Crime. Fine-tuning approaches use additional data for adaptation, which improves performance but requires substantial computational and data resources. VERA employs a verbalized learning structure \cite{xiao2024verbalized}, which requires a large amount of data (31,632 frames for UCF-Crime) and an additional VLM to generate interval verbal feedback. Nevertheless, CoReVAD achieves higher AP and AUC ROC than VERA on XD-Violence.
\FloatBarrier
\subsection{Ablation Study}
\label{sec:4.4}
In this section, we conduct ablation studies with UCF-Crime. First, we examine the effectiveness of the LRC. Next, we analyze the impact of the number of neighboring responses $l$. Finally, we assess the contribution of each component in the anomaly score refinement.
\begin{table}
\renewcommand{\arraystretch}{1.05}
\centering
\caption{Comparison of different cleaning strategies.}
\label{tab3}
\begin{tabular}{>{\arraybackslash}p{5cm} >{\centering\arraybackslash}p{1.5cm}}
\hline
\textbf{Cleaning}\par & \textbf{AUC(\%)}\par\\
\hline

No cleaning (baseline) & 81.51 \\
Caption cleaning \cite{zanella2024harnessing} & 80.47 \\
LRC & \textbf{82.51} \\
\hline
\end{tabular}
\end{table}

\noindent {\bfseries Effectiveness of Cleaning Strategies.} To validate the effectiveness of the Local Response Cleaning (LRC), we compare three strategies involving no cleaning, the cleaning method from LAVAD \cite{zanella2024harnessing}, and our LRC. As shown in Table~\ref{tab3}, we observe that CoReVAD achieves 81.51\% AUC ROC without any cleaning, but drops to 80.47\% when applying the conventional caption cleaning. As noted in Sect.~\ref{sec:3.2}, this performance drop results from the cleaning process yielding less relevant responses. In contrast, LRC achieves 82.51\%, which is about 2\% higher than the caption cleaning approach. This suggests that LRC mitigates noise while preserving local temporal context.
\begin{figure}
\centering
\includegraphics[width=0.7 \textwidth]{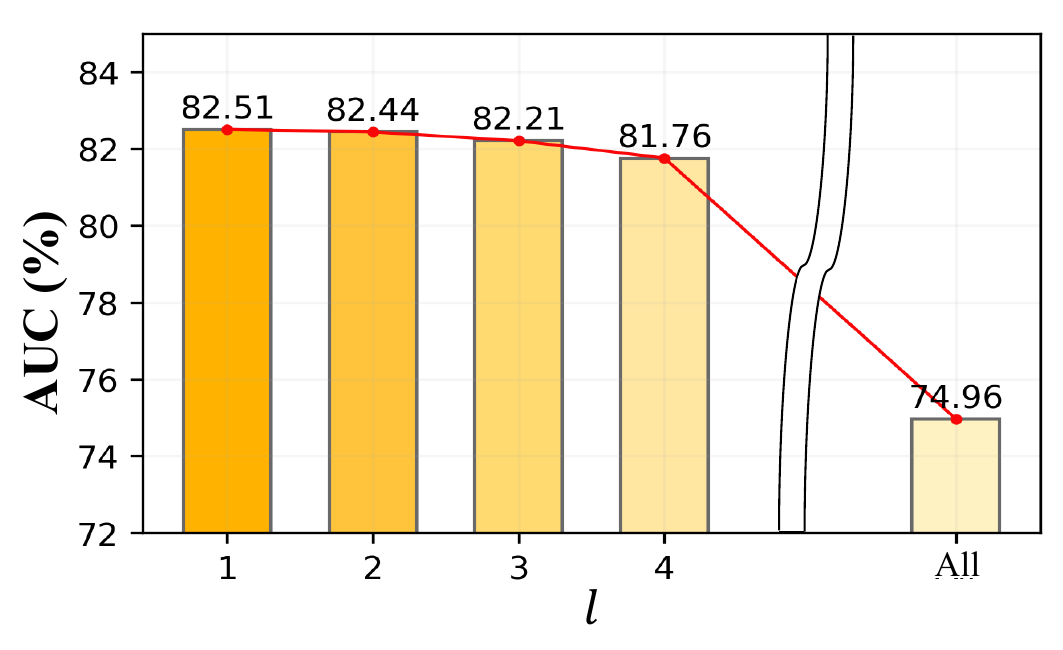}
\caption{Results of LRC over the number of $l$ neighboring segments used for reasoning selection.}
\label{fig4}
\end{figure}

\noindent{\bfseries Impact of neighboring responses.} We also analyze the impact of the number of neighboring responses $l$ used in LRC on VAD performance. As shown in Fig.~\ref{fig4}, increasing $l$ results in a gradual decrease in AUC ROC, reaching 74.96\% when all segments are considered. This suggests that incorporating a wider range of responses leads to the selection of contextually less relevant descriptions, which misaligns textual and visual semantics and reduces anomaly detection performance. In contrast, using a limited range of local context proves to be more effective for response cleaning.
\begin{table}
\renewcommand{\arraystretch}{1.05}
\centering
\caption{Effectiveness of each component for anomaly detection.}
\label{tab4}
\begin{tabular}{
    >{\centering\arraybackslash}m{3.2cm}
    >{\centering\arraybackslash}m{2.8cm}
    >{\centering\arraybackslash}m{2.8cm}
    >{\centering\arraybackslash}m{1.5cm}
}
\hline

\rule{0pt}{14pt}
\makecell[c]{\textbf{Vision-Text}\\\textbf{Score Refinement}} &
\makecell[c]{\textbf{Gaussian}\\\textbf{Smoothing}} &
\makecell[c]{\textbf{Position}\\\textbf{Weighting}} &
\makecell[c]{\textbf{AUC}\\\textbf{(\%)}} \\   
\hline

- & - & - & 73.76\\
\ding{51} & - & - & 81.46\\
\ding{51} & \ding{51} & - & 81.75\\
\ding{51} & \ding{51} & \ding{51} & \textbf{82.51}\\
\hline

\end{tabular}
\end{table}

\noindent {\bfseries Effectiveness of Each Component in the Anomaly Score Refinement.} To examine the impact of each refinement component in CoReVAD, we apply the components step by step and compare the results. As shown in Table~\ref{tab4}, using only the flattened initial scores without any refinement results in a relatively low performance of 73.76\% AUC ROC. Incorporating visual–text context for refinement substantially improves AUC ROC to 81.46\%, clearly demonstrating the effectiveness of this module. This result also shows that applying Gaussian smoothing and position weighting results in an improvement of 1.05\%.
\subsection{Qualitative Results}
\label{sec:4.5}
\begin{figure}
\centering
\includegraphics[width=\textwidth]{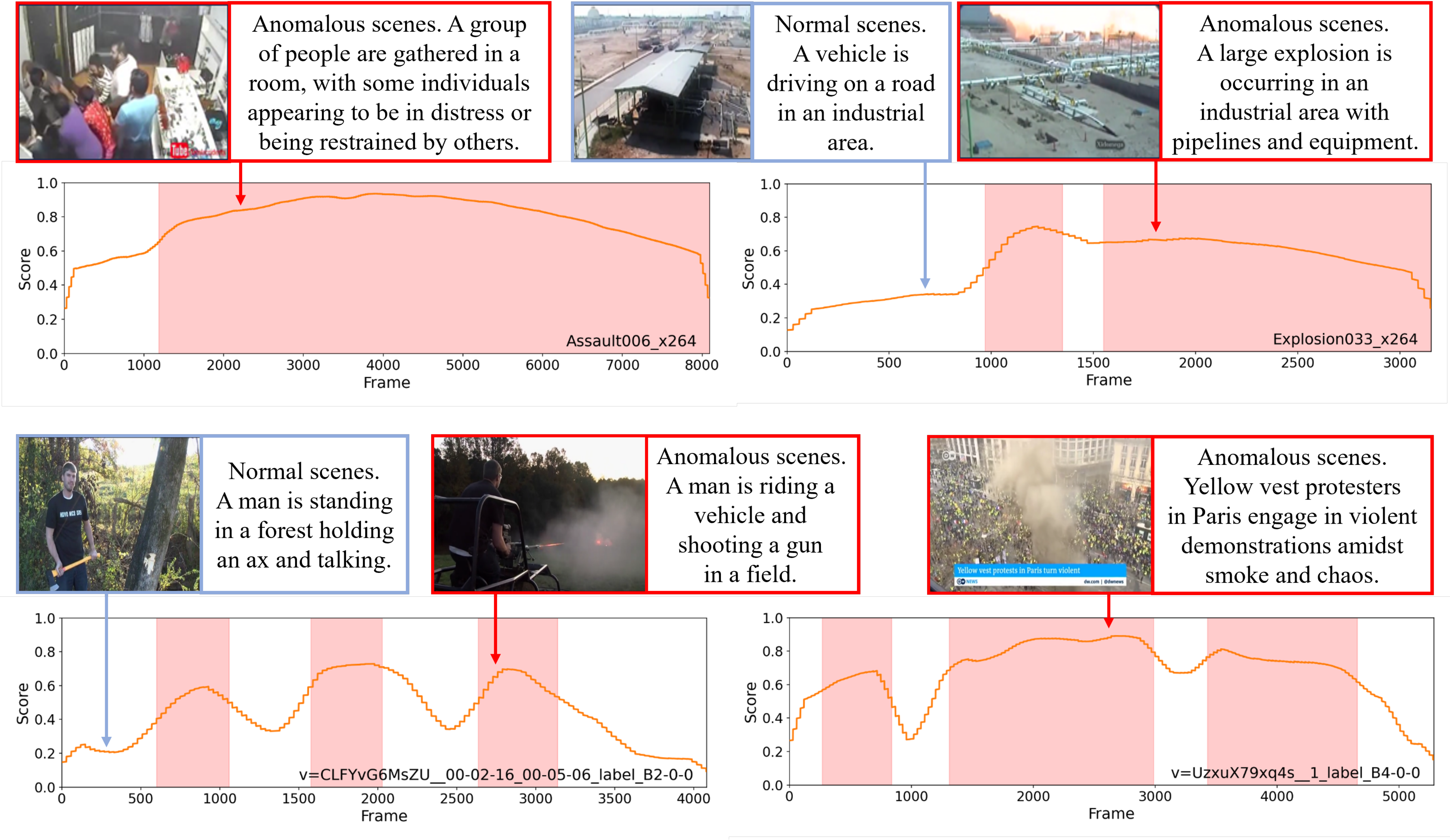}
\caption{Qualitative examples of the proposed method on UCF-Crime (top row) and XD-Violence (bottom row). We visualize the anomaly scores together with the temporal descriptions generated by our method. Blue boxes denote normal visual content and explanations, while red boxes denote abnormal scenes and their corresponding explanations. Ground-truth anomalies are highlighted in pink.}
\label{fig5}
\end{figure}

Fig.~\ref{fig5} presents the qualitative results of CoReVAD with sample videos from UCF-Crime and XD-Violence. The figure shows representative frames along with their corresponding temporal descriptions. In abnormal cases (red boxes), the model accurately describes visual content such as assaults, explosions, and shootings, which is well aligned with the high anomaly scores. Normal samples (blue boxes) also show low anomaly scores and generate descriptions that appropriately reflect the scene context. Overall, the qualitative examples demonstrate a clear alignment between the anomaly scores, explanations, and visual content. These results confirm the interpretability and robustness of our approach for explainable video anomaly detection. 
\section{Conclusion}
In this paper, we introduce CoReVAD, a training-free VAD framework that enables contextual reasoning using a single VLM. CoReVAD generates segment-level anomaly scores and explanations simultaneously, while the LRC module effectively reduces the randomness in VLM responses. In addition, an anomaly score refinement process that integrates both local and global temporal context enables precise frame-level anomaly scoring. Experimental results show that CoReVAD outperforms existing training-free approaches without requiring additional training or external LLMs, while also providing the interpretability expected in explainable VAD research. As with other VLM-based approaches, the performance of CoReVAD is influenced by the quality of the underlying pre-trained VLM and prompting strategy.


\begin{credits}
\subsubsection{\ackname}
This research was supported in part by Basic Science Research Program through the National Research Foundation of Korea (NRF) funded by the Ministry of Education (RS-2025-25415298), and by the Institute of Information and Communications Technology Planning and Evaluation (IITP) grant funded by the Korean Government through the Ministry of Science and ICT (MSIT) (XVoice: Multi-Modal Voice Meta Learning) under Grant 2022-0-00641.

\subsubsection{\discintname}
The authors have no competing interests to declare that are relevant to the content of this article.
\end{credits}
%
%
%
\bibliographystyle{splncs04}
\bibliography{ref}

\end{document}